\documentclass{ecai2004}
\usepackage{times}
\usepackage{graphicx}
\usepackage{latexsym}

\def\alcd{{\cal ALC}({\cal D})}
\def\csps{\mbox{\em CSPs}}
\def\tcsps{\mbox{\em TCSPs}}
\def\rcc8{\mbox{${\cal RCC}$8}}
\def\tcsp{\mbox{\em TCSP}}
\def\xdl{\mbox{\cal TCSP-ALC}({\cal D}_x)}
\def\cdalg{{{\cal CD}}}
\def\atra{{\cal CYC}_t}
\def\dnf{\mbox{\em DNF}}
\def\dmp{\mbox{\em DMP}}
\def\BBR{{\rm I\!R}}
\def\stp{\mbox{\em STP}}
\def\stps{\mbox{\em STPs}}
\def\allenb{<}
\def\allenm{\mbox{{\em m}}}
\def\alleno{\mbox{{\em o}}}
\def\allens{\mbox{{\em s}}}
\def\allend{\mbox{{\em d}}}
\def\allenf{\mbox{{\em f}}}
\def\allena{>}
\def\allenmi{\mbox{{\em mi}}}
\def\allenoi{\mbox{{\em oi}}}
\def\allensi{\mbox{{\em si}}}
\def\allendi{\mbox{{\em di}}}
\def\allenfi{\mbox{{\em fi}}}
\def\alleneq{\mbox{{\em eq}}}
\def\precedes{\mbox{\em PRECEDES}}
\def\intersects{\mbox{\em INTERSECTS}}
\def\follows{\mbox{\em FOLLOWS}}
\def\xdlzorcc{{\cal TCSP-ALC}({\cal D}_{\rcc8})}
\def\rccdc{{\mbox{\em DC}}}
\def\rccec{{\mbox{\em EC}}}
\def\rcctpp{{\mbox{\em TPP}}}
\def\rccpo{{\mbox{\em PO}}}
\def\rcceq{{\mbox{\em EQ}}}
\def\rccntpp{{\mbox{\em NTPP}}}
\def\rcctppi{{\mbox{\em TPPi}}}
\def\rccntppi{{\mbox{\em NTPPi}}}
\def\deuxdo{\mbox{2D}{\cal O}}
\def\apra{{\cal CYC}_b}
\def\rtopspace{{\cal RTS}}
\def\rccats{\mbox{$\rcc8${\em -at}}}
\def\atraats{\mbox{$\atra${\em -at}}}
\def\topspace{{\cal TS}}
\def\xat{{\mbox{x-{\em at}}}}
\def\alc{{\cal ALC}}
\def\wrt{\mbox{w.r.t.}}
\begin{document}

\title{\underline{\footnotesize{in Proceedings of the ECAI Workshop on Spatial and
	Temporal Reasoning, pp. 129-133, Valencia, Spain, 2004:}}\\
	An $\alcd$-based combination of temporal constraints and spatial
	constraints suitable for continuous (spatial) change}

\author{Amar Isli\\
FB Informatik, Universit\"at Hamburg\\
am99i@yahoo.com}

\maketitle
\bibliographystyle{ecai2004}
\underline{WORK EXACTLY AS REJECTED AT THE MAIN ECAI}\footnote{European
Conference on Artificial Intelligence.}\underline{ 2004}\footnote{The
\underline{reviews} are added to the actual paper, after the references,
for potential people interested in objectivity of conferences' reviewing
processes.}
\begin{abstract}
We present a family of spatio-temporal theories suitable for
continuous spatial change in general, and for continuous motion of
spatial scenes in particular. The family is obtained by
spatio-temporalising the well-known $\alcd$ family of Description
Logics (DLs) with a concrete domain D, as follows, where TCSPs
denotes "Temporal Constraint Satisfaction Problems", a well-known
constraint-based framework:
	(1) temporalisation of the roles, so that they consist of TCSP
	constraints (specifically, of an adaptation of TCSP
        constraints to interval variables); and
  	(2) spatialisation of the concrete domain $D$: the concrete
        domain is now $D_x$, and is  generated  by a spatial Relation
        Algebra (RA) $x$, in the style of the Region-Connection
        Calculus RCC8.
We assume durative truth (i.e., holding during a durative interval).
We also assume the homogeneity property  (if a truth holds during a
given interval, it holds during all of its subintervals). Among
other things, these assumptions raise the "conflicting" problem of
overlapping truths, which the work solves with the use of a specific
partition of the 13 atomic relations of Allen's interval algebra.

\end{abstract}
\newtheorem{cor}{Corollary}   
\newtheorem{rem}{Remark}
\newtheorem{thr}{Theorem}
\newtheorem{definition}{Definition}
\newtheorem{fact}{Fact}
\newtheorem{lem}{Lemma}
\newtheorem{example}{Example}
\section{Introduction}\label{sect1}
We start with our answer to the question of whether Artificial Intelligence (AI) should
reconsider, or revise its challenges:
\begin{flushright}
\begin{scriptsize}
``{\bf AI
to the service of the Earth as the Humanity's global,
continuous environment: the role of continuous (spatial) change in
building a lasting global, locally plausible democracy:}\\
(Cognitive) AI, which is guided by cognitively plausible assumptions on the physical world, such
as, e.g., {\em ``the continuity of (spatial) change''}, will start touching at its actual
success, the day it will have begun to serve, in return, as a source of inspiration for lasting
solutions to challenges such as, a World's globalisation respectful of local, regional beliefs
and traditions. One of the most urgent steps, we believe, is the implementation, in the Humanity's
global mind, of the idea of ``continuous change'', before any attempt of discontinuous
globalisation of our continuous Earth reaches a point of non return.''\\
\end{scriptsize}
\end{flushright}
Standard $\csps$ (Constraint Satisfaction Problems)
\cite{Mackworth77a,Montanari74a} were originally developed for
variables with discrete domains. With the aim of extending $\csps$ to
continuous variables, Dechter et al. \cite{Dechter91a} developed what 
is known in the literature as $\tcsps$ (Temporal Constraint
Satisfaction Problems), whose variables are continuous, in the sense
that they range over a continuous domain.

Constraint-based QSR (Qualitative Spatial Reasoning) languages very often consist of finite
RAs (Relation Algebras) \cite{Tarski41b}, with tables recording the results of applying
the different operations to the different atoms, and the reasoning issue
reduced to a matter of table look-ups: a good illustration to this is the
well-known topological calculus $\rcc8$ \cite{Randell92a} (see also \cite{Egenhofer91b}).

The goal of the present work is to combine $\tcsp$-like quantitative temporal
constraints with $\rcc8$-like qualitative spatial constraints. The targetted
applications are those involving motion, and spatial change in general, and
include reasoning about dynamic scenes in (high-level)
computer vision, and robot navigation. The framework we get can be seen as a spatio-temporalisation of
the well-known $\alcd$ family of Description Logics (DLs) with a concrete
domain ${\cal D}$ \cite{Baader91a}, which is
obtained by performing two specialisations at the same time:
(1) temporalisation of the roles, so that they consist of $\tcsp$
	constraints (specifically, of an adaptation of $\tcsp$ constraints to
	interval variables); and
(2) spatialisation of the concrete domain ${\cal D}$: the
    concrete domain is now ${\cal D}_x$, and is  generated  by a
    spatial Relation Algebra (RA) $x$, such as the Region-Connection Calculus RCC8
    \cite{Randell92a}.
The final spatio-temporalisation of $\alcd$ will be
referred to as $\xdl$, and its main properties can be summarised as follows:
(1) the (abstract) domain (i.e., the set of worlds in modal
    logics terminology) of $\xdl$ interpretations is a universe of
    time intervals;
(2) the roles consist of \mbox{4-argument} tuples providing
    $\tcsp$ constraints on the different pairs of endpoints of two
    intervals; and
(3) the concrete domain ${\cal D}_x$ is generated by an
    $\rcc8$-like constraint-based qualitative spatial language $x$.

Constraint-based languages candidate for generating a concrete domain
for a member of our family of spatio-temporal theories, are spatial RAs for
which the atomic relations form a decidable subset ---i.e., such
that consistency of a CSP expressed as a conjunction of $n$-ary
relations on $n$-tuples of objects, where $n$ is the arity of the RA
relations, is decidable. Examples of such RAs known in the literature
include
the Region-Connection Calculus $\rcc8$ in \cite{Randell92a} and
the projection-based Cardinal Direction Algebra $\cdalg$ in \cite{Frank92b},
for the binary case;
and
the RA $\atra$ of 2D orientations in \cite{Isli00b} for the
ternary case.

Examples of work in the literature on, or related to, change include
\cite{Halpern91a,Sandewall96b,Galton97a}. In this work, we are
interested in continuous change, and the approach we follow has
many similarities with the one in \cite{Halpern91a}. A first
difference with \cite{Halpern91a} is that, we will be interested
in representing continuous change, not only in propositional
truth (or knowledge), but in (relational) spatial truth as
well. Both truths hold during intervals. But contrary to the
approach in \cite{Halpern91a} (second difference), we consider
that truth is durative, in the sense that it holds during
durative, non-null intervals (intervals are thus interpreted as
in \cite{Allen83b}). An endpoint of an interval may or may not
belong to the interval (see, e.g., \cite{Galton97a} on this
issue, and on the issue of continuity in general).

The work can be seen as an extension of CSPs (Constraint
Satisfaction problems) of Allen's interval constraints
\cite{Allen83b}. An Allen's CSP is, in some sense, blind, in the
sense that, a solution to it is just a (consistent) collection
of intervals with the qualitative relation on each pair of them.
The solution tells nothing about possible change in truth (or
truths) in the real world. In this work, as mentioned above, we
consider two kinds of truth, propositional truth and
(relational) spatial truth. The first truth can be seen as a
propositonal formula, and the second as a similar formula, where,
instead of literals, we have qualitative spatial constraints on
objects of the spatial domain of interest (the domain of
intetrest may be a topological space, the objects
regions of that space, and the relations $\rcc8$ relations
\cite{Randell92a}). The conjunction of the two truths,
transformed into $\dnf$ (Disjunctive Normal Form), has disjuncts\footnote{A disjunct of a $\dnf$ is a
conjunction (of literals in the case of propositional calculus).}
consisting of conjuntions of literals and qualitative spatial
constraints. The $\dnf$ is true during an
interval, if one of its disjuncts is true during that interval. As a
consequence, if $C_1$ is true during interval $I_1$ and $C_2$
during interval $I_2$, and if $I_1$ and $I_2$ have a
1-dimensional intersection, then the conjunction of $C_1$ and
$C_2$ should be consistent. It follows that the following
partition of Allen's 13 atoms into three convex relations will be
primordial to our work:
(1) the relation consisting of the union of the before and meets atoms;
(2) its converse, containing the after and met-by relations;
and (3) the relation containing the remaining 9 atoms, which holds between two
intervals $\iff$ they have a 1-dimensional intersection.
Now, given truth $C_1$ holding during $I_1$ and truth $C_2$ holding
during $I_2$, $C_1$ and $C_2$ interact (i.e., their conjunction is
required to be consistent) $\iff$ $I_1$ and $I_2$ are related by the
third relation of the partition.

The paper, without loss of generality, will focus on a concrete
domain generated by one of the three binary spatial RAs mentioned above,
$\rcc8$ \cite{Randell92a}; and on another
concrete domain generated by the ternary spatial RA $\atra$ in
\cite{Isli00b}.
\section{Constraint satisfaction problems}\label{sect2}
A constraint satisfaction problem (CSP) of order $n$ consists of the following:
(1) a finite set of $n$ variables, $x_1,\ldots ,x_n$;
(2) a set $U$ (called the universe of the problem); and
(3) a set of constraints on values from $U$ which may be assigned to the variables.
The problem is solvable if
the constraints can be satisfied by some assignement of values
$a_1,\ldots ,a_n\in U$ to the variables $x_1,\ldots ,x_n$, in which case
the sequence $(a_1,\ldots ,a_n)$ is called a solution. Two problems are
equivalent if they have the same set of solutions.

An $m$-ary constraint is of the form $R(x_{i_1},\cdots ,x_{i_m})$, and asserts
that the $m$-tuple of values assigned to the variables $x_{i_1},\cdots ,x_{i_m}$
must lie in the $m$-ary relation $R$ (an $m$-ary relation over the
universe $U$ is any subset of $U^m$). An $m$-ary CSP is one of which the
constraints are $m$-ary constraints. We will be considering exclusively binary
CSPs and ternary CSPs.
\section{Temporal Constraint Satisfaction Problems ---$\tcsps$}
$\tcsps$ have been proposed in \cite{Dechter91a} as an extension of
(discrete) CSPs \cite{Mackworth77a,Montanari74a} to continuous variables.
\begin{definition}[$\tcsp$ \cite{Dechter91a}]
A $\tcsp$ consists of (1) a finite number of variables ranging over the
universe of time points; and (2) Dechter, Meiri and Pearl's
constraints (henceforth DMP constraints) on the variables.
\end{definition}
A $\dmp$ constraint is either unary or binary. A unary constraint has
the form $R(Y)$, and a binary constraint the form $R(X,Y)$, where $R$ is a subset of the set $\BBR$ of real
numbers, seen as a unary relation in the former case, and as a binary
relation in the latter case, and $X$ and $Y$ are variables ranging over
the universe of time points: the unary constraint $R(Y)$ is interpreted as $Y\in R$, and
the binary constraint $R(X,Y)$ as $(Y-X)\in R$. A unary constraint $R(Y)$ may 
be seen as a special binary constraint if we consider an origin of the 
World (time $0$), represented, say, by a variable $X_0$: $R(Y)$ is
then equivalent to $R(X_0,Y)$. Unless explicitly stated otherwise, we
assume, in the rest of the paper, that the constraints of a $\tcsp$
are all binary.
\begin{definition}[$\stp$ \cite{Dechter91a}]
An $\stp$ (Simple Temporal Problem) is a $\tcsp$ of which all the
constraints are convex, i.e., of the form $R(X,Y)$, $R$ being a convex 
subset of $\BBR$.
\end{definition}
The universal relation for $\tcsps$ in general, and for $\stps$ in
particular, is the relation consisting of the whole set $\BBR$ of real
numbers: the knowledge $(Y-X)\in\BBR$, expressed by the $\dmp$ constraint
$\BBR (X,Y)$, is equivalent to ``no knowledge''. The identity
relation is the (convex) set reducing to the singleton $\{0\}$: the
constraint $\{0\}(X,Y)$ ``forces" variables $X$ and $Y$ to be equal.
\section{A quick overview of Allen's interval algebra}
Allen's RA \cite{Allen83b} is well-known. Its importance for
this work is primordial, since it handles relations on temporal intervals, instead of
relations on temporal points as in the RA in \cite{Vilain86a}: as such, it captures much better
the idea of continuity of spatial change in the physical world \cite{Galton97a}. Briefly, the algebra is qualitative
and contains 13 atoms, which allow to differentiate between the 13 possible configurations
of two intervals on the time line. The atoms are $\allenb$ (before),
$\allenm$ (meets), $\alleno$ (overlaps), $\allens$ (starts), $\allend$ (during), $\allenf$
(finishes); their respective converses $\allena$ (after), $\allenmi$ (met-by), $\allenoi$
(overlapped-by), $\allensi$ (started-by), $\allendi$ (contains), $\allenfi$ (finished-by); and
$\alleneq$ (equals), which is its proper converse.

{\bf A partition suitable for continuous change.}
We will be using the partition of the set of Allen's atoms into three JEPD
(Jointly Exhaustive and Pairwise Disjoint) sets, which are $\precedes$, $\intersects$ and $\follows$, defined as follows:
$\precedes =\{\allenb ,\allenm\}$,
$\intersects =\{\alleno ,\allenoi ,\allens ,\allensi ,\allend ,\allendi ,\allenf ,\allenfi ,\alleneq\}$,
$\follows =\{\allenmi ,\allena\}$. The importance of this partition for
handling continuous (spatial, but also propositional) change will appear
later, but an intuitive explanation can be given right now. If a relation $r$
holds on a pair $(x,y)$ of spatial objects during interval $I$, then it holds
during all subintervals of $I$ (homogeneity property, see, e.g., \cite{Halpern91a}). In
the case of $r$ being disjunctive, we also assume that there exists an atom
$s$ in $r$ that holds on pair $(x,y)$ during interval $I$ ---without such an
additional assumption, if could be that an atom $s_1$ holds on $(x,y)$ during,
say, the first half of $I$, and another, distinct atom $s_2$ holds on $(x,y)$
during the other half of $I$. Given this property, if we have the knowledge
that (1) a relation $r_1$ holds on pair $(x,y)$ of spatial objects during
interval $I_1$; (2) a relation $r_2$ holds on the same pair during interval
$I_2$; and (3) intervals $I_1$ and $I_2$ are related by the relation
$\intersects$, then we conclude that relations $r_1$ and $r_2$ should have a
nonempty intersection (in particular, if they both consist of atomic relations,
they should be the same relation) ---one atom of $r_1\cap r_2$ holds then on
$(x,y)$ during $I_1\cup I_2$. This also applies to propositional knowledge.
If a propositional formula $\phi$ holds during interval $I$, then it holds
during all subintervals of $I$. In the case of $r$ being disjunctive, we also
assume that there exists a disjunct $c$ of the decomposition of $\phi$ into
$\dnf$ (Disjunctive Normal Form) such that $c$
holds during interval $I$
---here also, without such an additional assumption, if could be that a
disjunct $c_1$ holds during, say, the first half of $I$, and another disjunct
$c_2$, distinct from $c_1$, holds during the other half of $I$.
\section{A quick overview of the spatial relations to be used as the predicates
of the concrete domain}
\begin{figure*}
\centerline{\includegraphics{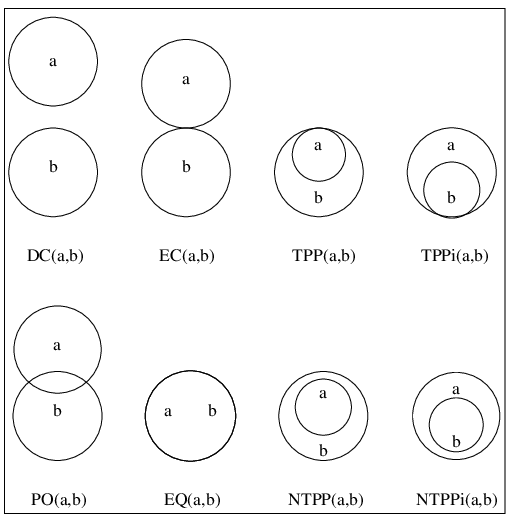}
	    \includegraphics{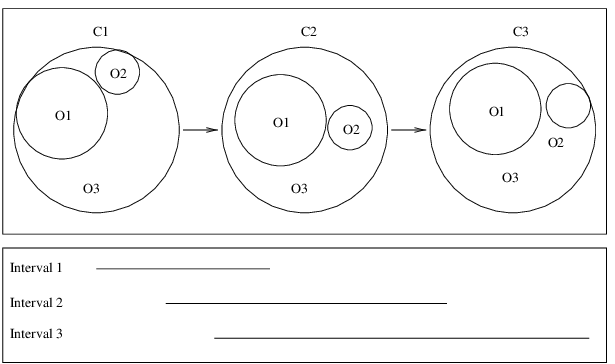}}
\caption{((Left) An illustration of the RCC-8 atoms. (Right) Illustration of $\xdlzorcc$.}\label{rcc-cda-atoms}
\end{figure*}
{\bf The RA $\rcc8$.}
The RCC-8 calculus (see \cite{Randell92a} for details) consists of a set of eight
JEPD atoms, $\rccdc$ (DisConnected), $\rccec$ (Externally Connected),
$\rcctpp$ (Tangential Proper Part), $\rccpo$ (Partial Overlap), $\rcceq$
(EQual),
$\rccntpp$ (Non Tangential Proper Part), and the converses, $\rcctppi$ and
$\rccntppi$, of $\rcctpp$ and $\rccntpp$, respectively.

{\bf The RA $\atra$.}
The set $\deuxdo$ of 2D orientations is defined in the usual way, and is isomorphic to the set of directed
lines incident with a fixed point, say $O$. Let $h$ be the natural isomorphism, associating with each
orientation $x$ the directed line (incident with $O$) of orientation $x$. The angle $\langle x,y\rangle$
between two orientations $x$ and $y$ is the anticlockwise angle $\langle h(x),h(y)\rangle$.
The binary RA of 2D orientations in \cite{Isli00b}, $\apra$, contains four atoms:
$e$ (equal), $l$ (left), $o$ (opposite) and $r$ (right). For all $x,y\in\deuxdo$:
$e(y,x) \Leftrightarrow \langle x,y\rangle =0$;
$l(y,x) \Leftrightarrow \langle x,y\rangle\in (0,\pi )$;
$o(y,x) \Leftrightarrow \langle x,y\rangle =\pi$;
$r(y,x) \Leftrightarrow \langle x,y\rangle\in (\pi ,2\pi )$.
Based on $\apra$, a ternary RA, $\atra$, for cyclic ordering of 2D
orientations has been defined in \cite{Isli00b}: $\atra$ has $24$ atoms, thus $2^{24}$
relations.
The atoms of $\atra$ are written as $b_1b_2b_3$, where
$b_1,b_2,b_3$ are atoms of $\apra$, and such an atom
is interpreted as follows:
$(\forall x,y,z\in\deuxdo )(b_1b_2b_3(x,y,z)\Leftrightarrow b_1(y,x)
 \wedge b_2(z,y)\wedge b_3(z,x))$.
The reader is referred to \cite{Isli00b} for more details.
\section{Concrete domain}
The role of a concrete domain in so-called DLs with a concrete domain \cite{Baader91a}, is to give
the user of the DL the opportunity to represent, thanks to predicates, knowledge on
objects of the application domain, as constraints on tuples of these objects.
\begin{definition}[concrete domain \cite{Baader91a}]\label{cddefinition}
A concrete domain ${\cal D}$ consists of a pair
$(\Delta _{{\cal D}},\Phi _{{\cal D}})$, where
$\Delta _{{\cal D}}$ is a set of (concrete) objects, and
$\Phi _{{\cal D}}$ is a set of predicates over the objects in $\Delta _{{\cal D}}$.
Each predicate $P\in\Phi _{{\cal D}}$ is associated
with an arity $n$: $P\subseteq (\Delta _{{\cal D}})^n$.
\end{definition}
\begin{definition}[admissibility \cite{Baader91a}]\label{cdadmissibility}
A concrete domain ${\cal D}$ is admissible if:
(1) the set of its predicates is closed under negation and
    contains a predicate for $\Delta _{{\cal D}}$; and
(2) the satisfiability problem for finite conjunctions of
    predicates is decidable.
\end{definition}
\section{The concrete domains ${\cal D}_x$,
            with $x\in\{\rcc8 ,\atra\}$}
The concrete domain generated by $x$, ${\cal D}_x$, can be written as
${\cal D}_x=(\Delta _{{\cal D}_x},\Phi _{{\cal D}_x})$, with
$                         {\cal D}_{\rcc8}   =(\rtopspace ,2^{\rccats})$ and
$                         {\cal D}_{\atra}   =(\deuxdo ,2^{\atraats})$,
where:
\begin{enumerate}
  \item $\rtopspace$ is the set of regions of a topological space $\topspace$;
    $\deuxdo$ is the set of 2D orientations; and
  \item $\xat$ is the set of $x$ atoms
    ---$2^{\xat}$ is thus the set of all $x$ relations.
\end{enumerate}
Admissibility of the concrete domains ${\cal D}_x$ is a direct consequence of
(decidability and) tractability of the subset $\{\{r\}|r\in\xat\}$ of $x$
atomic relations (see \cite{Renz99b} for
$x=\rcc8$, and \cite{Isli00b} for $x=\atra$).
\section{Syntax of $\xdl$ concepts,
            with $x\in\{\rcc8 ,\atra\}$}
Let $x$ be an RA from the set $\{\rcc8 ,\atra\}$. $\xdl$, as already explained,
is obtained from $\alcd$ by temporalising the roles, and spatialising the concrete domain.
The roles in $\alc$, as well as the roles other than the abstract features in $\alcd$, are
interpreted in a similar way as the modal operators of the multi-modal logic
${\cal K}_{(m)}$ \cite{Halpern85a} (${\cal K}_{(m)}$ is a multi-modal version of the
minimal normal modal system ${\cal K}$), which explains Schild's \cite{Schild91a}
correspondence between $\alc$ and ${\cal K}_{(m)}$. 
In this work, the roles will be
$4$-argument tuples, $\langle R^{bb},R^{be},R^{eb},R^{ee}\rangle$, with $R^{xy}$,
$x,y\in\{b,e\}$, being a convex subset of the set $\BBR$ of real numbers. The abstract
objects are intervals each of which is associated with (spatial) constraints on (objects
of) the scene of the spatial domain in consideration, and with propositional knowledge
consisting of primitive concepts and negated primitive concepts (literals): during the
whole interval, the scene has to fulfil the constraints (durativeness of spatial relational truth),
and the propositional knowledge has to remain true (durativeness of propositional truth). The
roles, thus, express temporal constraints on pairs of abstract
objects. Given an interval $I$, $I_b$ and $I_e$ will denote the
beginning endpoint of $I$ and the ending endpoint of $I$,
respectively. Given two intervals $I$ and $J$, $R^{xy}(I,J)$, with
$x,y\in\{b,e\}$, is interpreted as, the difference $J_y-I_x$,
representing the temporal distance between the endpoints $I_x$ and $J_y$, belongs
to the convex subset $R^{xy}$ of $\BBR$.

The assertion ``$I$ is an interval in the sense of Allen \cite{Allen83b} (a pair of temporal points such that the second strictly follows the first)'' can be expressed as $\langle \{0\},\BBR ^+,\BBR ^-,\{0\}\rangle (I,I)$. Now, given two intervals in the sense of Allen, and an Allen atom, say $r$, the constraint $r(I,J)$ is expressed as $s(I,J)$, where $s$ is the translation of $r$ as given by the following tables.
\begin{center}
$
\begin{array}{|l|l|}  \hline
Atom&Translation\\  \hline\hline
\allenb&\langle \BBR ^+,\BBR ^+,\BBR ^+,\BBR ^+\rangle\\ \hline
  \allenm&\langle \BBR ^+,\BBR ^+,\{0\},\BBR ^+\rangle\\ \hline
    \alleno&\langle \BBR ^+,\BBR ^+,\BBR ^-,\BBR ^+\rangle\\ \hline
      \allens&\langle \{0\},\BBR ^+,\BBR ^-,\BBR ^+\rangle\\ \hline
        \allend&\langle \BBR ^-,\BBR ^+,\BBR ^-,\BBR ^+\rangle\\ \hline
          \allenf&\langle \BBR ^-,\BBR ^+,\BBR ^-,\{0\}\rangle\\ \hline
            \allena&\langle \BBR ^-,\BBR ^-,\BBR ^-,\BBR ^-\rangle\\ \hline
\end{array}
\begin{array}{|l|l|}  \hline
Atom&Translation\\  \hline\hline
              \allenmi&\langle \BBR ^-,\{0\},\BBR ^-,\BBR ^-\rangle\\ \hline
                \allenoi&\langle \BBR ^-,\BBR ^+,\BBR ^-,\BBR ^+\rangle\\ \hline
                  \allensi&\langle \{0\},\BBR ^+,\BBR ^-,\BBR ^-\rangle\\ \hline
                    \allendi&\langle \BBR ^+,\BBR ^+,\BBR ^-,\BBR ^-\rangle\\ \hline
                      \allenfi&\langle \BBR ^+,\BBR ^+,\BBR ^-,\{0\}\rangle\\ \hline
                        \alleneq&\langle \{0\},\BBR ^+,\BBR ^+,\{0\}\rangle\\ \hline
\end{array}
$
\end{center}
The concepts of the $\xdl$
specialisation of the $\alcd$ family of DLs we will be interested in,
are built from three kinds of basic concepts:
\begin{enumerate}
  \item Primitive concepts (which play
    the role of atomic propositions in propositional calculus)
    and negated primitive concepts.
  \item Predicate concepts whose function can be described as
    follows. To describle an $\rcc8$-like spatial constraint of the form $P(x,y)$, where $P$ is a
    relation, and $x$ and $y$ variables, we use a predicate concept of the form
    $\exists (g_1)(g_2).P$, where $g_1$ and $g_2$ are concrete features
    referring, respectively, to the same concrete objects of the spatial 
    concrete domain in consideration as variables $x$ and
    $y$. Similarly, a ternary constraint of the form $P(x,y,z)$ will be
    represented by a predicate concept of the form $\exists
    (g_1)(g_2)(g_3).P$.
  \item The other basic concepts we will be using are of the form $\exists 
    R.A$, where $R=\langle R^{bb},R^{be},R^{eb},R^{ee}\rangle$ is a role and $A$ is a
    defined concept.
\end{enumerate}
Formally, the $\xdl$ concepts are defined as follows:
\begin{definition}[$\xdl$ concepts]\label{defxdlconcepts}
Let $x$ be an RA from the set $\{\rcc8 ,\atra\}$. Let $N_C$
and $N_{cF}$ be mutually disjoint and countably infinite sets of concept
names and concrete features, respectively. We suppose a partition
$N_C=N_{pC}\cup N_{dC}$ of $N_C$, where $N_{pC}$ is a set of primitive 
concepts, and $N_{dC}$ is a set of defined concepts. The set of $\xdl$
concepts is the smallest set such that:
\begin{enumerate}
  \item\label{defxdlconceptsone} $\top$ and $\bot$ are $\xdl$ concepts
  \item\label{defxdlconceptstwo} a $\xdl$ primitive concept is a $\xdl$
    (atomic) concept
  \item\label{defxdlconceptstwob} the negation, $\neg A$, of a $\xdl$
	primitive concept $A$ is a $\xdl$ concept
  \item\label{defxdlconceptsthree} if
    $A$ is a $\xdl$ defined concept;
    $C$ and $D$ are $\xdl$ concepts;
    $\langle R^{bb},R^{be},R^{eb},R^{ee}\rangle$ is a role;
    $g_1$, $g_2$ and $g_3$ are concrete features; and
    $P$ is a $\xdl$ predicate,
    then:
    \begin{enumerate}
      \item\label{defxdlconceptsthreeb}
            $\exists (g_1)(g_2).P$, if $x$ binary, and
            $\exists (g_1)(g_2)(g_3).P$, if $x$ ternary, are $\xdl$
            (atomic) concepts; and
      \item\label{defxdlconceptsthreea}
            $\exists\langle R^{bb},R^{be},R^{eb},R^{ee}\rangle .A$,
            $C\sqcap D$,
            $C\sqcup D$ are $\xdl$ concpets.
    \end{enumerate}
\end{enumerate}
\end{definition}
A ($\xdl$ terminological) axiom is an expression of the form $A\doteq C$, $A$ being
a defined concept and $C$ a concept. A TBox is a finite set of axioms, with
the condition that no defined concept appears more than once as the left hand side of
an axiom.
\begin{example}[illustration of $\xdlzorcc$]\label{xdlzorcc}Consider the moving spatial scene depicted in Figure
\ref{rcc-cda-atoms}(Right), composed of three objects o1, o2 and o3. Three snapshots of three submotions are presented, and
    associated with concepts $C_1$, $C_2$ and $C_3$. The configuration 
    described by the concept $C_1$ is so that, o1 is externally
    connected to o2, and
    is tangential proper part to o3, and the time interval during
    which the configuration holds overlaps the time interval during
    which the configuration described by the concept $C_2$ holds.
    The configuration described by the concept $C_2$ is so that, o1 is 
    externally connceted to o2, and
    o2 is non-tangential proper part to o3, and the time interval during
    which the configuration holds overlaps the time interval during
    which the configuration described by the concept $C_3$ holds.
    The configuration described by the concept $C_3$ is so that, o1 is 
    non-tangential proper part of o3, and the time interval during
    which the configuration holds is overlapped-by the time interval during
    which the configuration described by the concept $C_1$ holds.

We make use of
the concrete features $g_1$, $g_2$ and $g_3$ to refer to the actual regions corresponding to objects
o1, o2 and o3 in the scene. The TBox composed of the following axioms represents the
described moving spatial scene:
\begin{footnotesize}
\begin{eqnarray}
C_1&\doteq&\exists (g_{1})(g_{2}).\rccec\sqcap
           \exists (g_{1})(g_{3}).\rcctpp\sqcap
           \exists\langle\{0\},\BBR +,\BBR -,\{0\}\rangle .C_1\nonumber\\
   &      &\sqcap
           \exists\langle\BBR +,\BBR +,\BBR -,\BBR +\rangle .C_2
           \nonumber\\
C_2&\doteq&\exists (g_{1})(g_{2}).\rccec\sqcap
           \exists (g_{2})(g_{3}).\rccntpp\sqcap
           \exists\langle\{0\},\BBR +,\BBR -,\{0\}\rangle .C_2\nonumber\\
   &      &\sqcap
           \exists\langle\BBR +,\BBR +,\BBR -,\BBR +\rangle .C_3
           \nonumber\\
C_3&\doteq&\exists (g_{1})(g_{3}).\rccntpp\sqcap
           \exists\langle\{0\},\BBR +,\BBR -,\{0\}\rangle .C_3\nonumber\\
   &      &\sqcap
           \exists\langle\BBR -,\BBR +,\BBR -,\BBR -\rangle .C_1
           \nonumber
\end{eqnarray}
\end{footnotesize}
The situation described by the TBox is inconsistent for the following
reason. A defined concept describes a configuration of the spatial
scene which remains the same during the time interval associated with
the defined concept. As a consequence, if the time intervals
associated with two defined concepts overlap, the conjunction of the
correspondung two configurations of the scene should be
consistent. Concepts $C_1$ and $C_3$ in our example are so that, the
associated intervals overlap, but the conjunction $\exists (g_{1})(g_{2}).\rccec\sqcap
           \exists (g_{1})(g_{3}).\rcctpp\sqcap\exists (g_{1})(g_{3}).\rccntpp$ is not consistent.
\end{example}
\section{Semantics of $\xdl$,
            with $x\in\{\rcc8 ,\atra\}$}
As stated in the introduction, we intend our work to extend Allen's CSPs
\cite{Allen83b}, to make them ``see'' the reality of the physical world,
reality consisting, on the one hand, of (relational) spatial knowledge, recording,
e.g., the (durative) look of a spatial scene of interest at specific intervals
(spatial situation), and, on the other hand, of propositional knowledge, recording
the truth values of propositional variables at specific intervals (propositional
situation). We consider thus linear time, and a $\xdl$ interpretation will consist
of a collection of intervals of the time line, together with, on the one hand, a
truth assignment function, assigning with each interval the set of atomic
propositions true during that interval, and, on the other hand, of a finite number,
say n, of concrete features $g_1,\ldots ,g_n$, which are partial functions from the
set of intervals in the collection onto a universe of concrete spatial values.
Clearly, if such an interpretation is so that a given concrete feature is defined
for both of two intervals related by the $\intersects$ relation, then the value of
the concrete feature at one of the intervals should be the same as the one at the
other interval. Furthermore, if two intervals $I$ and $J$ are so that $I$ is a
subinterval of $J$ (i.e., related to it by the disjunctive relation $\{\allens ,
\alleneq ,\allend ,\allenf\}$) then the atomic propositions true during $J$ should
all be true during $I$ as well. Formally, a $\xdl$ interpretation is defined as
follows:
\begin{definition}[interpretation]
Let $x\in\{\rcc8 ,\atra\}$.
An interpretation ${\cal I}$ of $\xdl$ consists of a pair ${\cal I}=(t _{{\cal I}},.^{{\cal I}})$, where
$t _{{\cal I}}$ is a finite collection of intervals of the time line, and $.^{{\cal I}}$ is an
interpretation function mapping each primitive concept $A$ to a subset
$A^{{\cal I}}$ of $t _{{\cal I}}$, and each
concrete feature $g$ to a partial function $g^{{\cal I}}$:
\begin{footnotesize}
\begin{enumerate}
  \item from $t _{{\cal I}}$ onto the set $\rtopspace$ of regions of a topological space $\topspace$, if $x=\rcc8$;
  \item from $t _{{\cal I}}$ onto the set $\deuxdo$ of orientations of the 2-dimensional
    space, if $x=\atra$.
\end{enumerate}
\end{footnotesize}
\end{definition}
\begin{definition}[satisfiability of a TBox]
Let $x\in\{\rcc8 ,\atra\}$ be a spatial RA, ${\cal T}$ a $\xdl$
TBox, and ${\cal I}=(t_{{\cal I}},.^{{\cal I}})$ a
$\xdl$ interpretation. ${\cal T}$ is satisfiable by ${\cal I}$,
denoted ${\cal I}\models {\cal T}$, $\iff$ there exists a
one-to-one mapping $\phi$ from the set $t_{{\cal I}}$ of intervals
of ${\cal I}$ onto the set $D_{\cal T}$ of defined concepts appearing in
${\cal T}$, so that
${\cal I},s\models\langle\phi (s) ,{\cal T}\rangle$, for all
$s\in t_{{\cal I}}$. Satisfiability by $s\in t_{{\cal I}}$ of a concept
$C$ $\wrt$ ${\cal T}$, denoted by
${\cal I},s\models\langle C,{\cal T}\rangle$, is defined recursively as
follows:
\begin{enumerate}
  \item ${\cal I},s\models\langle B,{\cal T}\rangle$ $\iff$
    ${\cal I},s\models\langle C,{\cal T}\rangle$, for all defined
    concepts $B$ given by the axiom $B\doteq C$ of ${\cal T}$
  \item ${\cal I},s\models\langle\top ,{\cal T}\rangle$
  \item ${\cal I},s\not\models\langle\bot ,{\cal T}\rangle$
  \item ${\cal I},s\models\langle A,{\cal T}\rangle$ $\iff$
    $s\in A^{{\cal I}}$, and ${\cal I},s\models\langle\neg A,{\cal T}\rangle$
	$\iff$ ${\cal I},s\not\models\langle A,{\cal T}\rangle$, for all primitive concepts $A$
  \item ${\cal I},s\models\langle\exists (g_1)(g_2).P,{\cal T}\rangle$ $\iff$
    $P(g_1^{{\cal I}}(s),g_2^{{\cal I}}(s))$
  \item ${\cal I},s\models\langle\exists (g_1)(g_2)(g_3).P,{\cal T}\rangle$ $\iff$
    $P(g_1^{{\cal I}}(s),g_2^{{\cal I}}(s),g_3^{{\cal I}}(s))$
  \item ${\cal I},s\models\langle\exists\langle R^{bb},R^{be},R^{eb},R^{ee}\rangle .C,{\cal T}\rangle$ $\iff$
	$\langle R^{bb},R^{be},R^{eb},R^{ee}\rangle (s,\phi ^{-1}(C))$ and
	${\cal I},\phi ^{-1}(C)\models\langle C,{\cal T}\rangle$, for all defined concepts $C$
  \item ${\cal I},s\models\langle C\sqcap D,{\cal T}\rangle$ $\iff$
    ${\cal I},s\models\langle C,{\cal T}\rangle$ and ${\cal
      I},s\models\langle D,{\cal T}\rangle$
  \item ${\cal I},s\models\langle C\sqcup D,{\cal T}\rangle$ $\iff$
    ${\cal I},s\models\langle C,{\cal T}\rangle$ or ${\cal
      I},s\models\langle D,{\cal T}\rangle$
\end{enumerate}
\end{definition}
\section{Deciding satisfiability of a TBox ---an overview}
We describe briefly how to decide satisfiability of a TBox. We
suppose the particular case with all right hand sides of axioms
being conjunctions, and each role implying either of the three
relations in the already defined partition of Allen's atoms
into three disjunctive relations, $\precedes$, $\intersects$
and $\follows$. The general case  can be solved by
combining this particular case with recursive search. The idea
is to associate with the TBox a temporal CSP where the interval
variables are the defined concepts, and each subconcept of the
form $\exists R.D$ ($D$ is a defined concept), appearing in the
right hand side of the axiom defining a defined concept $C$,
giving rise to the constraint $R(C,D)$. The second step is to
associate with each of the temporal variables a conjunction
consisting of one subconjunction which is a propositional formula,
and another subconjunction which is a spatial CSP expressed in
the RA $x$. Furthermore, if two interval variables are related by
the $\intersects$ relation, then the propositional-spatial
conjunction associated with one is augmented with that associated
with the other (so that the homogeneity property gets satisfied).
Decidability now is a consequence of decidability of a proposition
formula, of a spatial CSP expressed in the RA $x$, and of a
temporal CSP involving only $\precedes$, $\intersects$ and
$\follows$ (and the universal relation) ---which can solved
polynomially by translating it into the convex part of $\tcsps$
\cite{Dechter91a}, known as $\stps$.
\section{Summary}
We have presented an $\alcd$-based combination of temporal
constraints and spatial constraints suitable for the representation
of continuous change in the real physical world. The approach
handles both spatial and propositional change. Knowledge about
continuous change is represented as a TBox, and we have shown that
satisfiability of such a TBox is decidable.
\bibliography{/home/AmarUn/Research/BIBLIO/amar}
\begin{center}
THE NOTIFICATION LETTER\\
(as received on 3 May 2004)
\end{center}
Dear Amar Isli:

We regret to inform you that your submission 

  C0693
  An ALC(D)-based combination of temporal constraints and spatial
  constraints suitable for continuous (spatial) change -first results
  Amar Isli
 
cannot be accepted for inclusion in the ECAI 2004's programme. Due to 
the large number of submitted papers, we are aware that also otherwise 
worthwhile papers had to be excluded. You may then consider submitting 
your contribution to one of the ECAI's workshops, which are still open 
for submission.

In this letter you will find enclosed the referees' comments on your 
paper.

We would very much appreciate your participation in the meeting and 
especially in the discussions. 

Please have a look at the ECAI 2004 website for registration details 
and up-to-date information on workshops and tutorials:

    http://www.dsic.upv.es/ecai2004/

The schedule of the conference sessions will be available in May 2004. 

I thanks you again for submitting to ECAI 2004 and look forward to 
meeting you in Valencia.

Best regards
      
Programme Committee Chair
\begin{center}
REVIEW ONE
\end{center}
----- ECAI 2004 REVIEW SHEET FOR AUTHORS -----

PAPER NR: C0693 

TITLE: An ALC(D)-based combination of temporal constraints and spatial 
constraints suitable for continuous (spatial) change

1) SUMMARY (please provide brief answers)

- What is/are the main contribution(s) of the paper?

The paper describes the integration of quantitative representations of 
temporal constraints with qualitative representations of spacial 
constraints within a framework of ALC(D) Description Logics.

2) TYPE OF THE PAPER

The paper reports on:

  [X] Preliminary research

  [ ] Mature research, but work still in progress

  [ ] Completed research

The emphasis of the paper is on:

  [ ] Applications

  [X] Methodology

3) GENERAL RATINGS

Please rate the 6 following criteria by, each time, using only 
one of the five following words: BAD, WEAK, FAIR, GOOD, EXCELLENT

3a) Relevance to ECAI: FAIR

3b) Originality: FAIR

3c) Significance, Usefulness: FAIR

3d) Technical soundness: FAIR

3e) References: WEAK

3f) Presentation: BAD

4) QUALITY OF RESEARCH

4a) Is the research technically sound?

    [ ] Yes   [X] Somewhat   [ ] No

4b) Are technical limitations/difficulties adequately discussed?

    [ ] Yes   [X] Somewhat   [ ] No

4c) Is the approach adequately evaluated?

    [ ] Yes   [ ] Somewhat   [X] No

FOR PAPERS FOCUSING ON APPLICATIONS:

4d) Is the application domain adequately described?

    [ ] Yes   [ ] Somewhat   [ ] No

4e) Is the choice of a particular methodology discussed?

    [ ] Yes   [ ] Somewhat   [ ] No

FOR PAPERS DESCRIBING A METHODOLOGY:

4f) Is the methodology adequately described?

    [ ] Yes   [X] Somewhat   [ ] No

4g) Is the application range of the methodology adequately described, 
    e.g. through clear examples of its usage?

    [ ] Yes   [ ] Somewhat   [X] No

Comments:

The quality of presentation of the paper is not sufficient to make a 
reliable judgment regarding the general quality of the research, hence 
the largely neutral ratings of this section.

5) PRESENTATION

5a) Are the title and abstract appropriate? 

    [ ] Yes   [ ] Somewhat   [X] No

5b) Is the paper well-organized? [ ] Yes   [ ] Somewhat   [X] No

5c) Is the paper easy to read and understand? 

    [ ] Yes   [ ] Somewhat   [X] No

5d) Are figures/tables/illustrations sufficient? 

    [ ] Yes   [ ] Somewhat   [X] No

5e) The English is  [ ] very good   [ ] acceptable   [X] dreadful

5f) Is the paper free of typographical/grammatical errors? 

    [ ] Yes   [ ] Somewhat   [X] No

5g) Is the references section complete?

    [ ] Yes   [X] Somewhat   [ ] No

Comments:

The presentation of this work lets it down completely. It is below the 
standard necessary for a general international audience of AI 
researchers, and this virtually debars it from the possibility of a measured 
technical evaluation. The paper tries to cram far too much technical 
detail into too little space, at the expense of any high-level, informal or 
intuitive description of the work, or any detailed indication of its 
applicability. The one example in the paper is badly described, and comes 
too late to help readability. The grammar is in many places tortuously 
long-winded and over-complex, uses commas gratuitously, and in some 
places is virtually unparsable (e.g. the first sentence of the fourth 
paragraph of the introduction). The unassigned, utterly pretentious and 
absurd quotation at the beginning of the introduction is particularly 
unhelpful and inappropriate.

6) TECHNICAL ASPECTS TO BE DISCUSSED (detailed comments)

- Suggested / required modifications:

To be acceptable for publication within the given page limitation, the 
paper needs to be restructured, and a different presentational style 
needs to be adopted. At the level of overall structure, I suggest that 
the authors describe their work at least partially with the aid of 
illustrative, running examples and/or application-based problems. At a more 
detailed level, I suggest adopting a much more concise grammatical 
style.

- Other comments:

\begin{center}
REVIEW TWO
\end{center}
----- ECAI 2004 REVIEW SHEET FOR AUTHORS -----

PAPER NR: C0693 

TITLE: An ALC(D) based combination of temporal and spatil constraints

1) SUMMARY (please provide brief answers)

- What is/are the main contribution(s) of the paper?

The authors try to combine Allen's interval algebra, Dechter et al's 
TCSP, Randell et al's RCC8 and Isli's Cyc\_t into an ALC(D) framework for 
dealing with spatio-temporal change.

2) TYPE OF THE PAPER

The paper reports on:

  [X] Preliminary research

  [ ] Mature research, but work still in progress

  [ ] Completed research

The emphasis of the paper is on:

  [ ] Applications

  [X] Methodology

3) GENERAL RATINGS

Please rate the 6 following criteria by, each time, using only 
one of the five following words: BAD, WEAK, FAIR, GOOD, EXCELLENT

3a) Relevance to ECAI: GOOD

3b) Originality: FAIR

3c) Significance, Usefulness: WEAK

3d) Technical soundness: FAIR

3e) References: FAIR

3f) Presentation: WEAK

4) QUALITY OF RESEARCH

4a) Is the research technically sound?

    [ ] Yes   [X] Somewhat   [ ] No

4b) Are technical limitations/difficulties adequately discussed?

    [ ] Yes   [ ] Somewhat   [X] No

4c) Is the approach adequately evaluated?

    [ ] Yes   [ ] Somewhat   [X] No

FOR PAPERS FOCUSING ON APPLICATIONS:

4d) Is the application domain adequately described?

    [ ] Yes   [ ] Somewhat   [ ] No

4e) Is the choice of a particular methodology discussed?

    [ ] Yes   [ ] Somewhat   [ ] No

FOR PAPERS DESCRIBING A METHODOLOGY:

4f) Is the methodology adequately described?

    [ ] Yes   [ ] Somewhat   [X] No

4g) Is the application range of the methodology adequately described, 
    e.g. through clear examples of its usage?

    [ ] Yes   [ ] Somewhat   [X] No

Comments:

It is tried to put too many things together without giving sufficient 
motivation why it is done and what is the use of it. The chosen calculi 
have nothing to do with each other and the combination seems completely 
arbitrary. E.g. how does Cyc\_t fit into it and why is it chosen. Any 
other calculus could have been chosen as well. 
All in all it is much too dense and too confusing for readers being 
able to extract the main ideas of the paper. I suggest that either it is 
tried to restrict to a combination of only two calculi first or to write 
a longer version (with substantial motivation and explanation) and 
submit it to a journal.

5) PRESENTATION

5a) Are the title and abstract appropriate? 

    [ ] Yes   [X] Somewhat   [ ] No

5b) Is the paper well-organized? [ ] Yes   [ ] Somewhat   [X] No

5c) Is the paper easy to read and understand? 

    [ ] Yes   [ ] Somewhat   [X] No

5d) Are figures/tables/illustrations sufficient? 

    [ ] Yes   [ ] Somewhat   [X] No

5e) The English is  [ ] very good   [X] acceptable   [ ] dreadful

5f) Is the paper free of typographical/grammatical errors? 

    [ ] Yes   [X] Somewhat   [ ] No

5g) Is the references section complete?

    [ ] Yes   [X] Somewhat   [ ] No

Comments:

Maybe add a reference to Gerevini and Nebel's ECAI 2002 paper on the 
combination of interval and RCC8 relations.

6) TECHNICAL ASPECTS TO BE DISCUSSED (detailed comments)

- Suggested / required modifications:

Sections 3 and 4 could be made shorter and more precise. 

Do you consider all 13 interval relations or just the three you 
introduce in section 4? 

Figure 1. How is it possible that C1, C2 (and C3) hold at the same time 
(the intervals during which they hold overlap)?

- Other comments:

What is "RCC8-like"? 

Section 8, the {e,b} notation should be explained. I guess it means 
beginning and end point?

\begin{center}
REVIEW THREE
\end{center}
----- ECAI 2004 REVIEW SHEET FOR AUTHORS -----

PAPER NR: C0693 

TITLE: An ALC(D)-based combination of temporal constraints and 
spatial...

1) SUMMARY (please provide brief answers)

- What is/are the main contribution(s) of the paper?

It's hard to find a contribution in this paper.

2) TYPE OF THE PAPER

The paper reports on:

  [X] Preliminary research

  [ ] Mature research, but work still in progress

  [ ] Completed research

The emphasis of the paper is on:

  [ ] Applications

  [X] Methodology

3) GENERAL RATINGS

Please rate the 6 following criteria by, each time, using only 
one of the five following words: BAD, WEAK, FAIR, GOOD, EXCELLENT

3a) Relevance to ECAI: WEAK

3b) Originality: BAD

3c) Significance, Usefulness: BAD

3d) Technical soundness: BAD

3e) References: BAD

3f) Presentation: BAD

4) QUALITY OF RESEARCH

4a) Is the research technically sound?

    [ ] Yes   [ ] Somewhat   [X] No

4b) Are technical limitations/difficulties adequately discussed?

    [ ] Yes   [ ] Somewhat   [ ] No

4c) Is the approach adequately evaluated?

    [ ] Yes   [ ] Somewhat   [X] No

FOR PAPERS FOCUSING ON APPLICATIONS:

4d) Is the application domain adequately described?

    [ ] Yes   [ ] Somewhat   [ ] No

4e) Is the choice of a particular methodology discussed?

    [ ] Yes   [ ] Somewhat   [ ] No

FOR PAPERS DESCRIBING A METHODOLOGY:

4f) Is the methodology adequately described?

    [ ] Yes   [ ] Somewhat   [X] No

4g) Is the application range of the methodology adequately described, 
    e.g. through clear examples of its usage?

    [ ] Yes   [ ] Somewhat   [X] No

Comments:

5) PRESENTATION

5a) Are the title and abstract appropriate? 

    [ ] Yes   [ ] Somewhat   [X] No

5b) Is the paper well-organized? [ ] Yes   [ ] Somewhat   [X] No

5c) Is the paper easy to read and understand? 

    [ ] Yes   [ ] Somewhat   [X] No

5d) Are figures/tables/illustrations sufficient? 

    [ ] Yes   [X] Somewhat   [ ] No

5e) The English is  [ ] very good   [X] acceptable   [ ] dreadful

5f) Is the paper free of typographical/grammatical errors? 

    [ ] Yes   [X] Somewhat   [ ] No

5g) Is the references section complete?

    [ ] Yes   [ ] Somewhat   [X] No

Comments:

6) TECHNICAL ASPECTS TO BE DISCUSSED (detailed comments)

- Suggested / required modifications:

This is a very confusing paper in many respects.

1. It is claimed to develop a Spatio-Temporal description logic 
starting from the concrete domain approach ALC(D). Anyway, there is no 
evidence the author understood that a DL MUST have an object domain while, as 
far as I understood, the semantics is based on a combination of spatial 
and temporal domain. Thus, the semantics is based ONLY on a concrete 
domains and there is no mention of the abstract object domain. DLs are 
mainly formalism for representing and reasoning about domain objects. 
Then, for sure, you can EXTEND them by introducing other concrete domains. 
Indeed, in the pseudo-DL presented here roles are JUST Allen temporal 
relations.

2. The Pseudo-DL is not even ALC since there is NO full negation but 
just primitive negation. Indeed, using full negation the logic would be a 
sub-case of the undecidable Halper and Shoham interval modal logic.

3. Homogeneity. First of all, the reference of Halpern and Shoham has 
nothing to do with such property. There are many papers studying such 
property (e.g., Allen, Shoham, etc.) but they are not mentioned. 
Furthermore, the claimed homogeneity is not reflected by the semantic presented 
in Section 9.

4. The Semantic is hard to understand. Definition 7 makes few sense. 
Furthermore, the temporal domain used (finite collection of intervals) is 
not interesting at all. Temporal structures should be based on Natural, 
Real or Rational numbers.

5. Related works. It's quite amazing that the author(s) disregard the 
literature on temporal and spatial DLs. I would like to mention that 
many papers appeared in the literature in the last 10 years but there is 
no mention of them AT ALL! Just to mention few authors: Artale-Franconi, 
Bettini, Schmiedel (interval based DLs); Baader, Lutz (complexity 
results on various concrete domain extensions); Zachariashev-Wolter, Schild 
(point-based DLs); Zachariashev-Wolter (spatial extensions of DL's).

- Other comments:
\end{document}